\documentclass[sigconf]{acmart}

\usepackage{booktabs}
\usepackage{graphicx}
\usepackage{amsmath}
\usepackage{threeparttable}

\AtBeginDocument{%
  \providecommand\BibTeX{{%
    \normalfont B\kern-0.5em{\scshape i\kern-0.25em b}\kern-0.8em\TeX}}}

\setcopyright{acmlicensed}
\copyrightyear{2018}
\acmYear{2018}
\acmDOI{XXXXXXX.XXXXXXX}

\acmConference[MM'24]{Make sure to enter the correct
  conference title from your rights confirmation email}{October 28 - November 1,
  2024}{Melbourne, Australia.}
\acmISBN{978-1-4503-XXXX-X/18/06}

\begin{document}

\title{LVLM\_CSP: Accelerating Large Vision Language Models via Clustering, Scattering, and Pruning for Reasoning Segmentation}


\author{Hanning Chen}
\affiliation{%
  \institution{University of California, Irvine}
  \city{Irvine, CA}
  \country{USA}}
\email{hanningc@uci.edu}

\author{Yang Ni}
\affiliation{%
  \institution{University of California, Irvine}
  \city{Irvine, CA}
  \country{USA}}
\email{yni3@uci.edu}

\author{Wenjun Huang}
\affiliation{%
  \institution{University of California, Irvine}
  \city{Irvine, CA}
  \country{USA}}
\email{wenjunh3@uci.edu}

\author{Hyunwoo Oh}
\affiliation{%
  \institution{University of California, Irvine}
  \city{Irvine, CA}
  \country{USA}}
\email{hyunwooo@uci.edu}

\author{Yezi Liu}
\affiliation{%
  \institution{University of California, Irvine}
  \city{Irvine, CA}
  \country{USA}}
\email{yezil3@uci.edu}

\author{Tamoghno Das}
\affiliation{%
  \institution{University of California, Irvine}
  \city{Irvine, CA}
  \country{USA}}
\email{tamoghnd@uci.edu}

\author{Mohsen Imani}
\affiliation{%
  \institution{University of California, Irvine}
  \city{Irvine, CA}
  \country{USA}}
\email{m.imani@uci.edu}

\renewcommand{\shortauthors}{author name and author name, et al.}

\begin{abstract}
  Large Vision Language Models (LVLMs) have been widely adopted to guide vision foundation models in performing reasoning segmentation tasks, achieving impressive performance. However, the substantial computational overhead associated with LVLMs presents a new challenge. The primary source of this computational cost arises from processing hundreds of image tokens. Therefore, an effective strategy to mitigate such overhead is to reduce the number of image tokens—a process known as image token pruning. Previous studies on image token pruning for LVLMs have primarily focused on high-level visual understanding tasks, such as visual question answering and image captioning. In contrast, guiding vision foundation models to generate accurate visual masks based on textual queries demands precise semantic and spatial reasoning capabilities. Consequently, pruning methods must carefully control individual image tokens throughout the LVLM reasoning process. Our empirical analysis reveals that existing methods struggle to adequately balance reductions in computational overhead with the necessity to maintain high segmentation accuracy. In this work, we propose \textbf{LVLM\_CSP}, a novel training-free visual token pruning method specifically designed for LVLM-based reasoning segmentation tasks. \textbf{LVLM\_CSP} consists of three stages: clustering, scattering, and pruning. Initially, the LVLM performs coarse-grained visual reasoning using a subset of selected image tokens. Next, fine-grained reasoning is conducted, and finally, most visual tokens are pruned in the last stage. Extensive experiments demonstrate that \textbf{LVLM\_CSP} achieves a \textbf{65\%} reduction in image token inference FLOPs with virtually no accuracy degradation, and a \textbf{70\%} reduction with only a minor \textbf{1\%} drop in accuracy on the 7B LVLM.
\end{abstract}





\maketitle

\section{Introduction}
\label{sec:intro}

\begin{figure}[t]
  \centering \includegraphics[width=\columnwidth]{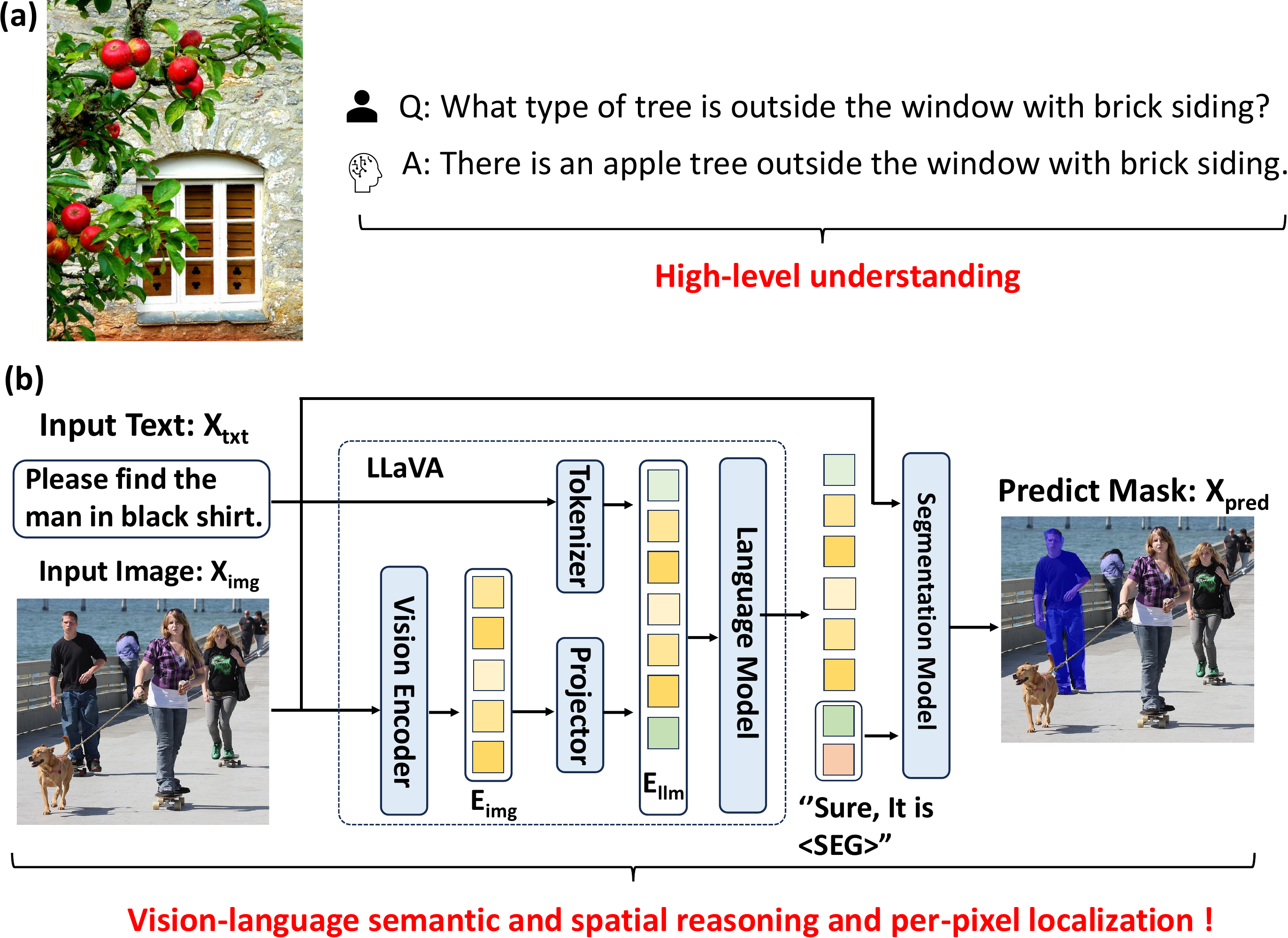}
   \caption{(a) An example of high-level image understanding. (b) LVLM-guided reasoning segmentation.}
   \label{fig:task_motivation}
\end{figure}

Large Vision-Language Models (LVLMs)~\cite{liu2023visual,zhu2023minigpt,li2023blip} have demonstrated strong capabilities in visual reasoning tasks, such as image understanding and visual question answering (VQA). Recently, several works~\cite{lai2024lisa,zhang2024groundhog,qian2024affordancellm,wei2024hyperseg,rasheed2024glamm}  
have proposed integrating LVLMs with vision foundation models (i.e., segmentation models)~\cite{kirillov2023segment,liu2024grounding} to address reasoning-based segmentation tasks, such as RefCOCO~\cite{yu2016modeling}, ReasoningSeg~\cite{lai2024lisa}, and RIO~\cite{qu2023rio}, achieving promising results. Unlike image understanding tasks, where LVLMs only need to provide high-level image information (as illustrated in Figure~\ref{fig:task_motivation}(a)), reasoning-based segmentation tasks require LVLMs to perform fine-grained reasoning about both semantic relationships (i.e., identifying which instance in the image matches the query text) and spatial relationships (i.e., determining the spatial location of the target instance) among the query text and multiple objects. Additionally, LVLMs must guide segmentation models to produce high-quality masks, as shown in Figure~\ref{fig:task_motivation}(b). Compared to previous vision-language models, such as MDETR~\cite{kamath2021mdetr} and SegCLIP~\cite{luo2023segclip}, LVLMs' robust visual reasoning capabilities enable the overall system to achieve significantly higher segmentation accuracy, particularly when addressing more complex and indirect tasks~\cite{chen2024vltp}.

However, LVLMs also introduce additional computational overhead. Among all computational components, the multi-head self-attention (MHSA)~\cite{vaswani2017attention} in the decoder of large language models (LLMs), when applied to visual tokens (i.e., image tokens), constitutes the majority. Taking LLaVA~\cite{liu2023visual} as an example, when using CLIP ViT-L with an input resolution of 224$\times$224, there are 256 image tokens participating in LLM decoder MHSA computations. Using CLIP ViT-L with a resolution of 336$\times$336 further increases the number to 576 image tokens. In both cases, there are significantly more image tokens than LLMs' system tokens and user-provided text query tokens~\cite{touvron2023llama}. Therefore, one of the most effective methods for accelerating LVLMs is to reduce the number of image tokens, also known as visual token pruning~\cite{tang2024survey}.

Although several prior works~\cite{chen2024image,shang2024llava,yang2024visionzip,zhang2024sparsevlm,xing2024pyramiddrop} have attempted to reduce image tokens in LVLMs for image understanding tasks, none have evaluated their effectiveness on reasoning-based segmentation tasks. As our empirical study demonstrates later in Section~\ref{sec:motivation}, previous methods are unable to effectively compress LVLMs for these challenging tasks due to their stringent requirements on image token control. Therefore, in this paper, we propose \textbf{LVLM\_CSP}, a large vision-language model pruning framework specifically designed for reasoning-based segmentation tasks. LVLM\_CSP adopts a three-stage approach: first, it enables LVLMs to perform \textbf{coarse-grained} visual reasoning on a carefully selected small set of image tokens in the clustering stage; second, it reactivates all image tokens, allowing LVLMs to carry out detailed \textbf{fine-grained} visual reasoning in the scattering stage; finally, in the pruning stage, having established a comprehensive understanding of relationships between query text and image instances, the framework aggressively drops image tokens, retaining only a minimal subset for subsequent computations. Consequently, the total computational cost of the LVLM is significantly reduced. Unlike previous methods, which either reduce image tokens at a single early stage~\cite{chen2024image,shang2024llava} or gradually remove tokens~\cite{xing2024pyramiddrop,endo2024feather}, our three-stage pruning framework effectively balances token compression with the ability of LVLMs to perform thorough visual reasoning across all image tokens. The contributions of the work are summarized as follows:

\begin{itemize}
\item We propose a three-stage visual token pruning framework for LVLMs designed for reasoning-based segmentation tasks. Our method, consisting of an initial coarse-grained stage, followed by a fine-grained stage, and concluding with an aggressive token-dropping stage, achieves an effective balance between model compression and complex visual reasoning.

\item To the best of our knowledge, this is the first LVLM visual token pruning method tailored for reasoning-based segmentation tasks. Compared to previous works, our method provides more precise control over image tokens.

\item Experiments conducted on various reasoning-based segmentation tasks demonstrate that LVLM\_CSP can reduce LVLM computations by 65\% without compromising accuracy, and can further reduce computations by 70\% with only a marginal accuracy drop of 1\%.
\end{itemize}

\section{Preliminary}

\subsection{LVLM in Reasoning Segmentation}
Figure~\ref{fig:task_motivation}.(b) illustrates a general framework for reasoning-based segmentation with LVLMs~\cite{lai2024lisa,qian2024affordancellm}. The LVLM reasons over image and text inputs to produce a guidance token (e.g., [SEG]), which is then used by a segmentation model (e.g., SAM~\cite{kirillov2023segment} or GroundingDINO~\cite{liu2024grounding}) to generate the object mask. An LVLM typically comprises a vision encoder (e.g., CLIP~\cite{radford2021learning}) and a language model (e.g., LLaMA~\cite{touvron2023llama}). The vision encoder transforms the image $\textbf{X}_\textbf{img}$ into visual token embeddings $\textbf{E}_{\textbf{img}} \in \mathbb{R}^{B \times N \times d}$, where $B$, $N$, and $d$ denote batch size, number of tokens, and embedding dimension. These embeddings are projected and concatenated with text embeddings. In LLaVA~\cite{liu2023visual}, the final input to the LLM is:
\begin{equation} \label{eq:llm_concate}
\mathbf{E_{llm}} = \mathbf{E_{sys}} | \mathbf{E_{img}} | \mathbf{E_{usr}}
\end{equation}
where $\textbf{E}_\textbf{sys}$ and $\textbf{E}_\textbf{usr}$ are system and user embeddings. $\textbf{E}_\textbf{llm}$ is processed by MHSA layers~\cite{vaswani2017attention} to produce the guidance token, which is passed to the segmentation model to generate the final mask.

\subsection{The Challenge of Existing Methods} \label{sec:motivation}

\begin{figure}[t]
  \centering \includegraphics[width=0.8\linewidth]{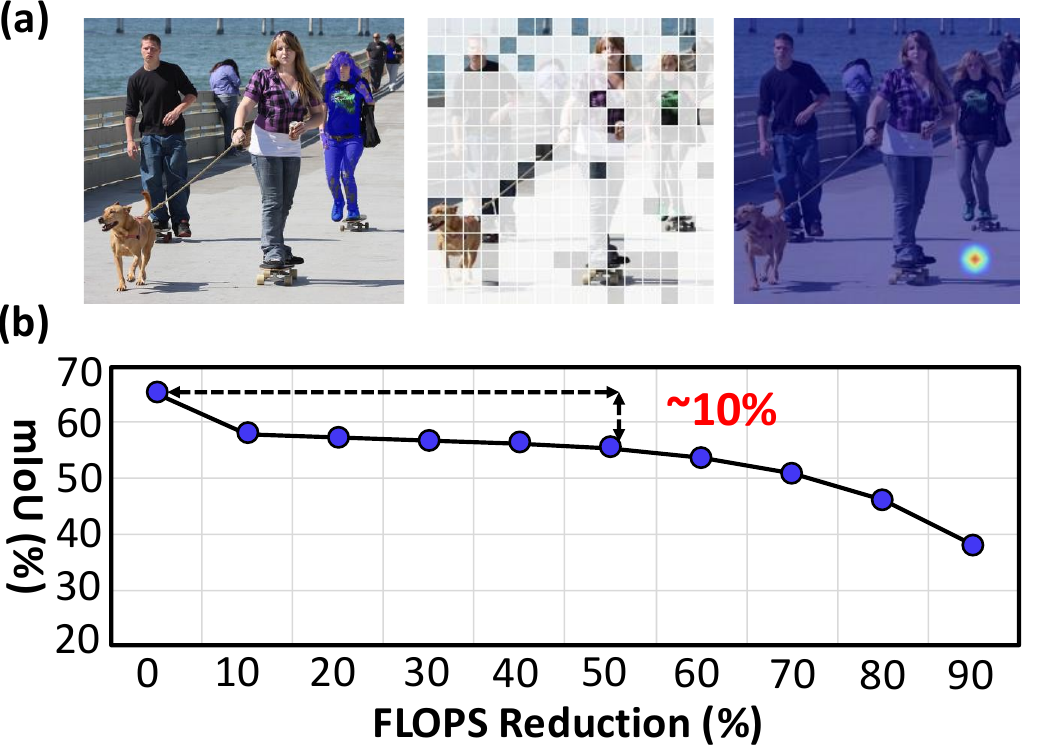}
   \caption{(a) Visualization of the final reasoning segmentation mask, token dropping, and attention map using $\textbf{LLaVA-PruMerge+}$~\cite{shang2024llava}. (b) The tradeoff curve between efficiency (FLOPS reduction ratio) and accuracy (mIoU) for $\textbf{LLaVA-PruMerge+}$.}
   \label{fig:PrueMerge_motivation}
\end{figure}

\begin{figure}[t]
  \centering \includegraphics[width=0.9\linewidth]{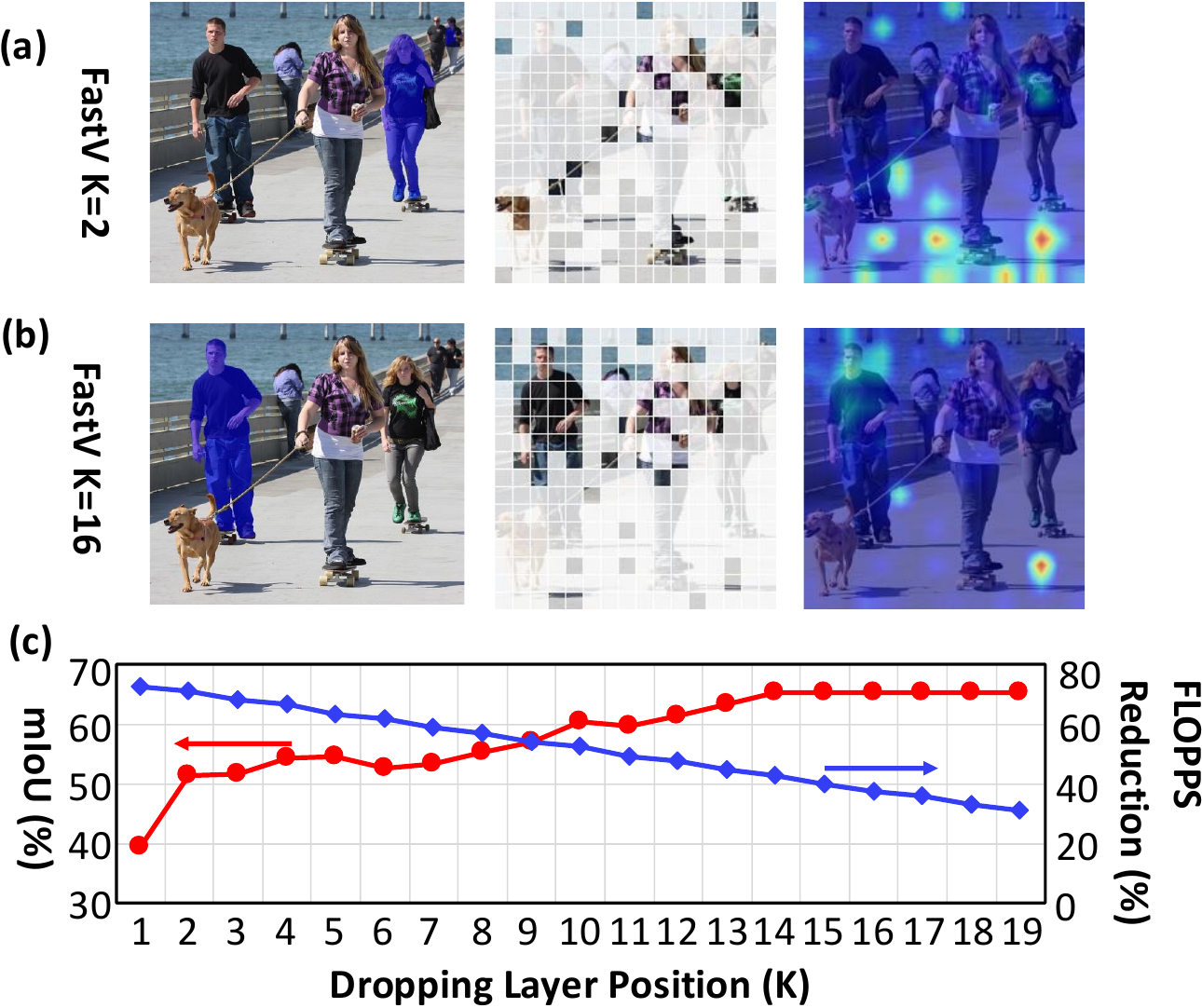}
   \caption{(a) Visualization of the final reasoning segmentation mask, token dropping, and attention map using $\textbf{FastV}$ \cite{chen2024image}, with the token dropping layer position set to $K=2$ and the number of retained visual tokens set to 64. (b) Same visualization for FastV with drop layer at $K=16$. (c) Trade-off curve between efficiency (FLOPs reduction ratio) and accuracy (mIoU) for $\textbf{FastV}$ when tuning the dropping layer position K.}
   \label{fig:FastV_motivation}
\end{figure}

Existing LVLM token reduction methods can generally be categorized into two types: (1) reducing visual tokens before the LLM, as in LLaVA-PruMerge~\cite{shang2024llava} (PruMerge), and (2) reducing image tokens within the LLM, as in FastV~\cite{chen2024image}. Both approaches primarily focus on visual token reduction for high-level visual understanding tasks such as image comprehension and VQA~\cite{liu2024improved}. However, when applied to reasoning-based segmentation tasks like RefCOCO, both methods exhibit limitations and fail to achieve optimal performance.

Figure~\ref{fig:PrueMerge_motivation} illustrates the impact of applying PruMerge to RefCOCO. The query text, identical to that in Figure~\ref{fig:task_motivation}(b), asks: “Please find the man in the black shirt.” PruMerge attempts to merge image tokens with similar semantic information. However, as shown in Figure~\ref{fig:PrueMerge_motivation}(b), dropping half of the image tokens results in approximately a 10\% decrease in mIoU. Furthermore, in Figure~\ref{fig:PrueMerge_motivation}(a), the LVLM provides incorrect guidance to the segmentation model, failing to retain the most important tokens, which leads to an inaccurate final mask. \textbf{The challenge of pruning before the LLM lies in the coarse-grained semantic representation of the vision encoder.} Directly merging image tokens reduces the granularity of visual information, making it difficult for the LLM to perform accurate reasoning.


Figure~\ref{fig:FastV_motivation} shows the effect of applying FastV to the same task. FastV drops visual tokens with low attention to output tokens. However, as shown in Figure~\ref{fig:FastV_motivation}(c), early token dropping significantly reduces mIoU, since the LLM has not yet grasped the query–image relationship. This leads to incorrect token selection and segmentation (Figure~\ref{fig:FastV_motivation}(a)). In contrast, dropping after sufficient MHSA layers (e.g., 16) allows the model to reason more accurately, improving both outcomes (Figure~\ref{fig:FastV_motivation}(b)). However, delaying token dropping reduces computational savings (Figure~\ref{fig:FastV_motivation}(c)), \textbf{highlighting a trade-off between accuracy and efficiency.} Therefore, existing LVLM image tokens reduction methods are difficult to apply directly to reasoning segmentation tasks. To address this, we propose LVLM\_CSP.

\section{LVLM\_CSP Design}

\begin{figure*}[t]
  \centering \includegraphics[width=\textwidth]{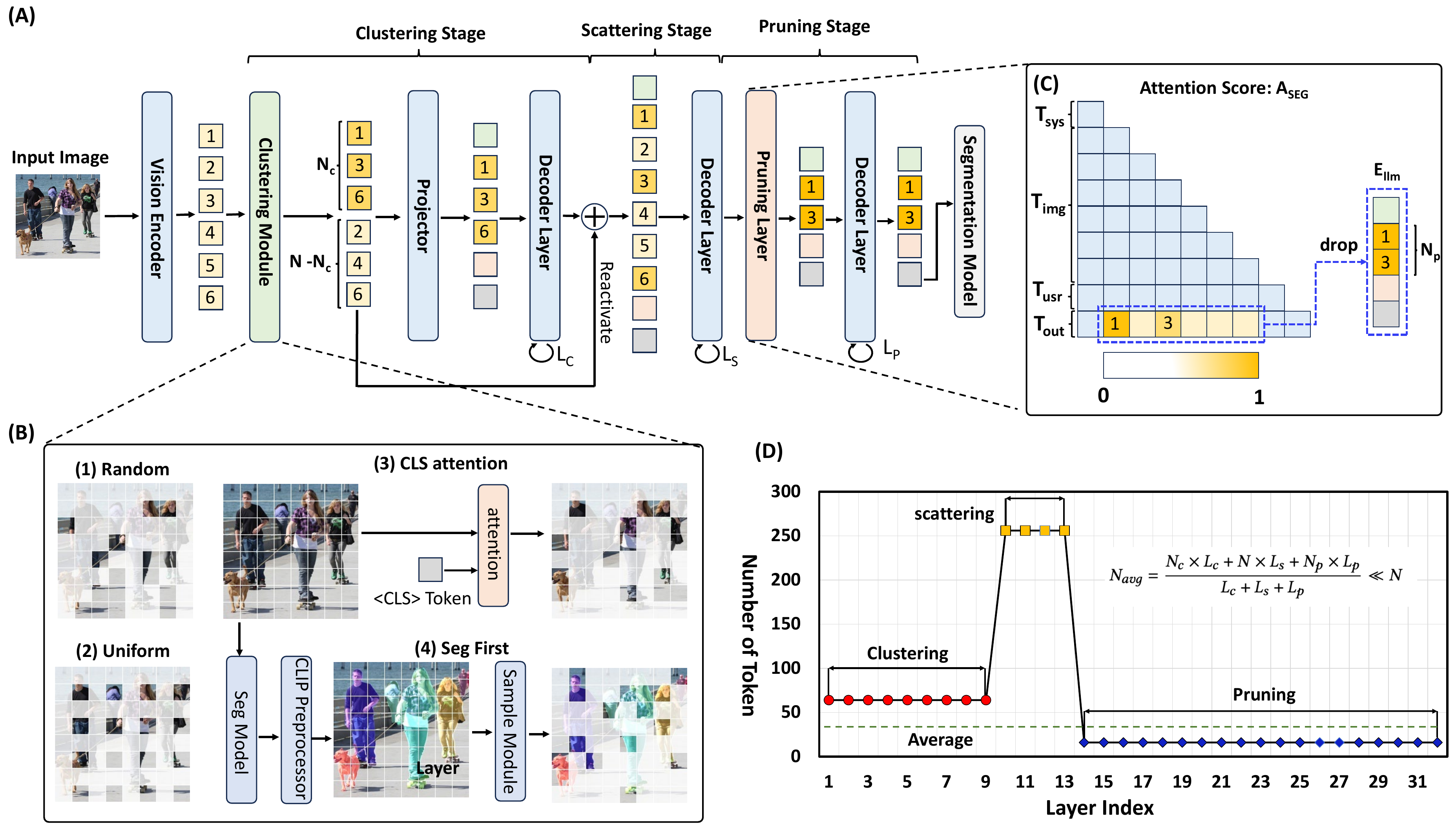}
  \caption{(A) Overview of LVLM\_CSP. (B) Design of the clustering module. (C) Design of the pruning layer. (D) Number of tokens participating in MHSA at each layer.}
  \label{fig:CSP_arch}
\end{figure*}

In this section, we introduce LVLM\_CSP, a LVLM pruning framework targeting reasoning segmentation tasks. LVLM\_CSP includes three stages, including \textbf{clustering stage}, \textbf{scattering stage}, and \textbf{pruning stage}, as is shown in Figure~\ref{fig:CSP_arch}.

\subsection{Clustering Stage}

During the clustering stage, we partition the image feature map into multiple clusters and select one representative token per cluster to feed into the LLM decoder’s attention layers. Unlike prior work~\cite{shang2024llava}, which fuses all image tokens and relies on the LLM decoder for direct fine-grained reasoning, our approach emphasizes coarse-grained reasoning. At this stage, the LLM only needs to form a high-level understanding of the task, while fine-grained reasoning is deferred to later stages.

Let the image token embedding $\textbf{E}_\textbf{img} \in \mathbb{R}^{N \times d}$, where $N$ is the number of image tokens produced by the vision encoder (CLIP ViT~\cite{liu2023visual}) and $d$ is the embedding dimension. We insert the clustering module ($\textbf{F}\textbf{c}$) between the vision encoder and the projection layer. $\textbf{F}\textbf{c}$ groups the image tokens into clusters and selects $\textbf{N}_\textbf{c}$ representative tokens, which are passed to the projection layer and the LLM decoder for coarse reasoning. The remaining $N - N_c$ tokens are not discarded—they are reintroduced in the scattering stage. Figure~\ref{fig:CSP_arch}.(B)(1–4) shows four different designs of the clustering module.

\noindent \textbf{Random} Figure~\ref{fig:CSP_arch}.(B)(1) illustrates the Random strategy, where $N_c$ image tokens are randomly selected from the total $N$ image tokens. 

\noindent \textbf{Uniform} Figure~\ref{fig:CSP_arch}.(B)(2) depicts the Uniform strategy. Let \( \textbf{S} \) denote the set of sampled token indices in the original image embedding, where \( \textbf{S[i]} \) represents the index of the \( i^{th} \) sampled token. The indices are determined as follows:
\begin{equation}
    \mathbf{S[i]} =  k \times \lfloor \frac{N}{N_c} \rfloor, \quad i = 0, 1, 2, \dots, N_c - 1.
\end{equation}
Uniform clustering ensures that the LLM reasons over the entire image. However, it may select redundant background tokens (e.g., floor in Figure~\ref{fig:CSP_arch}.(B)(2)) and still sample uninformative regions when objects are clustered on one side. To mitigate this, we propose the CLS attention and Seg-first methods.

\noindent \textbf{CLS Attention} The vision encoder of LLaVA~\cite{liu2023visual} typically utilizes the CLIP ViT model~\cite{radford2021learning}. In this model, in addition to processing image patch tokens, a class token ([CLS] token) is computed after the ViT self-attention layers to aggregate global image information. In the CLS Attention strategy, we use the attention scores between the [CLS] token and other visual tokens to determine the importance of each visual token. 
Let $\vec{q_{cls}}$ denote the query embedding of the [CLS] token from the last layer of CLIP ViT, and let $K_{img}$ represent the visual token key embeddings from the last layer, where $d$ is the embedding dimension. The clustering index is computed as:
\begin{equation}
    \mathbf{S} = \text{TopK}\left(\text{softmax}\left(\frac{q_{cls} \cdot K_{img}}{\sqrt{d}}\right), N_c\right)
\end{equation}
The [CLS] token captures high-level global features, often attending to large dominant objects while overlooking smaller ones~\cite{gao2024clip}. For instance, in Figure~\ref{fig:CSP_arch}.(B)(3), it gives low attention to the man in the black shirt, selecting only one image token for representation.

\noindent \textbf{Seg First} To address the challenge where the [CLS] token may overlook small object in the image, we propose the Segmentation First (Seg First) design, as is shown in Figure~\ref{fig:CSP_arch}.(B)(4). In this approach, we first utilize a lightweight segmentation model, such as Yolo-tiny~\cite{hussain2023yolo}, to perform instance-level segmentation and generate $N_o$ instance masks \textbf{M} with a shape of $N_o \times (H' \times W')$, where $H'$ and $W'$ correspond to the original dimensions of the input image. Next, we employ the CLIP image preprocessor to reshape the instance mask into $N_o \times (H \times W)$, where $H \times W$ is the CLIP ViT input image resolution. Within the sampler module, we sample image tokens from each instance’s corresponding token set based on its relative size:

\begin{equation}
    \mathbf{S_o[i]} = \max\left(1, \left\lfloor N_c \times \frac{\sum_{k=0}^{H \times W - 1} M[i,k]}{\sum_{j=0}^{N_o - 1} \sum_{k=0}^{H \times W - 1} M[j,k]} \right\rfloor \right)
\end{equation}
Here, $S_o[i]$ represents the number of visual tokens that need to be sampled for the $i^{th}$ object mask. We ensure that even for very small objects, at least one token is selected. The final selected token indices are determined as follows:
\begin{equation} \label{eq:seg_first_sample}
    \mathbf{S} = \bigcup_{i=1}^{N_o} s \sim \text{Uniform} \left( \binom{M[i]}{S_o[i]} \right)
\end{equation}
Equation~\ref{eq:seg_first_sample} means the sampler module will random select $S_o[i]$ image tokens from instance i. The final selcted tokens \textbf{S} is the union of all instances' selected tokens. As illustrated in Figure~\ref{fig:CSP_arch}.(B)(4), we guarantee that each object is sampled with at least one token. This enables the Large Language Model (LLM) to perform coarse-grained reasoning over all visual objects. The computational overhead of the tiny segmentation model is negligible. For instance, Yolo-tiny requires only 10 GFLOPs, which amounts to merely $\sim$0.1\% of the computation required by Llama-7B.

After the clustering module, we pass both the $N_c$ selected image token embeddings and the remaining $N - N_c$ image token embeddings to the projector. However, during the clustering stage, only the $N_c$ selected image tokens are concatenated with system tokens and user tokens before being fed into the LLM decoder. To preserve relative spatial information and prepare for fine-grained reasoning in the later scattering stage, we apply Rotary Position Embedding~\cite{su2024roformer} (RoPE) to the $N_c$ image tokens while maintaining their original position IDs (if no pruning is applied). Let $\textbf{P}_\textbf{img}$ be the position IDs of image tokens with a shape of $N_c \times 1$. Then, we define:

\begin{equation}
    P_{img}[i] = N_{sys} + S[i], \quad i \in [0, N_c - 1]
\end{equation}

where $N_{sys}$ denotes the number of system tokens preceding the image tokens. The same principle is also applied to user tokens and generated [SEG] tokens. 
These embeddings ($\textbf{E}_\textbf{llm}$) are then concatenated (equation~\ref{eq:llm_concate}) and processed through $\textbf{L}_\textbf{C}$ LLM decoder layers for coarse-grained reasoning.

\subsection{Scattering Stage}
After the coarse-grained attention between the [SEG] token and the \(N_c\) selected image tokens, the [SEG] token gains a general intuition about the direction for solving the query. To obtain more precise spatial and semantic information about the object, we perform fine-grained reasoning at the scattering stage. Specifically, we re-activate the \(N - N_c\) image tokens that were not selected during the clustering stage. 
The updated image embedding \(E_{img}^{S}\) is defined as follows:

\begin{equation}
    \mathbf{E_{img}^{S}}[i] = \begin{cases} 
E_{img}^{C}[i], & \text{if } i\in \mathbf{S} \\
E_{img}[i], & \text{otherwise}
\end{cases}
\end{equation}

The shape of the current image embedding is \(N \times D\). Due to the coarse-grained reasoning, the number of LLM MHSA layers at the scattering stage is lower than that at the clustering stage. That is, in Figure~\ref{fig:CSP_arch}.(A), we have \(L_S < L_C\). 

\subsection{Pruning Stage}
After fine-grained reasoning, we can now use the attention between the [SEG] token ($T_{out}$) and image tokens ($T_{img}$) to identify and retain the most relevant image tokens while discarding most of the irrelevant ones. Unlike previous works~\cite{chen2024image}, which utilize the average attention score of all generated tokens, since LVLM\_CSP focuses on reasoning-based segmentation tasks, we consider only the [SEG] token's attention score, as is shown in Figure~\ref{fig:CSP_arch}.(C). Thus, we define:
\begin{equation}
    \mathbf{A}_{SEG} = \mathbf{A}[I_{seg}][N_{sys}:N_{sys}+N]
\end{equation}
Here $I_{SEG}$ is the index of the seg token in the whole $\textbf{E}_\textbf{llm}$. For the remaining \(L_{P}\) MHSA layers, we select the top \(N_{P}\) image tokens based on the attention scores \(\mathbf{A}_{SEG}\). Since, after both coarse-grained and fine-grained attention, the LLM has already performed relatively accurate semantic and spatial reasoning, the pruning stage requires retaining only a very small fraction of image tokens, i.e., \(N_P \ll N\), as shown in Figure~\ref{fig:CSP_arch}.(A). 

\subsection{Computation Analysis}
As shown in Figure~\ref{fig:CSP_arch}.(D), unlike previous works~\cite{shang2024llava,chen2024image,xing2024pyramiddrop,endo2024feather} that either prune image tokens only once or drop them incrementally, LVLM\_{CSP} follows a three-stage process. It first removes image tokens for coarse-grained reasoning, then increases the number of image tokens for fine-grained reasoning, and finally applies precise and aggressive image tokens drop at pruning stage. The average number of image tokens ($\textbf{N}_{\textbf{avg}}$) participating in each LLM MHSA layer is given by:
\begin{equation}
    N_{avg} = \frac{N_c\times L_c + N\times L_s + N_p\times L_p}{L_c + L_s + L_p} \ll N
\end{equation}

\begin{table*}[]
\centering
\caption{Comparison of LVLM\_CSP with previous LVLM visual token pruning methods on LLaVA-1.5-7B, using Llama2 7B as the language model and \textbf{CLIP ViT-L patch14-224} as the vision encoder, at a pruning ratio of 70\%, evaluated across five reasoning segmentation datasets.}
\label{tab:lisa7b_224}
\resizebox{\textwidth}{!}{%
\begin{tabular}{c|ccc|ccc|cc|cc|cc|c|cc}
\toprule
 &
  \multicolumn{3}{c|}{\textbf{RefCOCO}} &
  \multicolumn{3}{c|}{\textbf{RefCOCO+}} &
  \multicolumn{2}{c|}{\textbf{RefCOCOg}} &
  \multicolumn{2}{c|}{\textbf{ReasonSeg}} &
  \multicolumn{2}{c|}{\textbf{RIO}} &
  $\textbf{mIoU}_\textbf{drop}$ &
  $\textbf{N}_\textbf{avg}$ &
  \textbf{TFLOPs} \\ \midrule
            & val  & testA & testB & val  & testA & testB & val  & test & val  & test & common & uncommon &      &     &      \\ \midrule
LLaVA-1.5-7B~\cite{lai2024lisa}     & 72.8 & 75.4  & 68.7  & 60.9 & 65.2  & 53.9  & 65.9 & 66.8 & 97.1 & 49.2 & 60.1   & 36.9     & 0.0 $\downarrow$ & 256 & 4.12 \\ \midrule
PruMerge~\cite{shang2024llava}    & 62.6 & 64.7  & 60.4  & 47.3 & 50.8  & 42.9  & 53.6 & 53.7 & 89.9 & 42.4 & 55.6   & 31.9     & 9.8 $\downarrow$ & 75  & 1.79 \\
PruMerge+~\cite{shang2024llava}   & 65.4 & 68.4  & 62.1  & 50.7 & 54.9  & 44.4  & 56.3 & 56.5 & 92.3 & 44.9 & 56.2   & 33.4     & 7.3 $\downarrow$ & 95  & 2.04 \\
VisionZip~\cite{yang2024visionzip}   & 64.6 & 67.6  & 62.3  & 49.7 & 53.5  & 44.4  & 56.5 & 56.6 & 91.3 & 43.1 & 56.1   & 33.7     & 7.8 $\downarrow$ & 80  & 1.85 \\ \midrule
FastV~\cite{chen2024image}       & 64.4 & 66.4  & 60.8  & 49.1 & 51.7  & 44.2  & 54.9 & 55.3 & 82.3 & 36.6 & 54.3   & 32.1     & 10.1 $\downarrow$ & 76  & 1.8  \\
Simignore~\cite{zhang2024enhancing}   & 64.4 & 66.3  & 60.4  & 49.3 & 51.4  & 44.1  & 54.6 & 55.1 & 82.5 & 36.3 & 54.1   & 31.9     & 10.2 $\downarrow$ & 76  & 1.8  \\
SparseVLM~\cite{zhang2024sparsevlm}   & 64.3 & 66.5  & 60.9  & 49.7 & 53.5  & 44.4  & 56.6 & 56.6 & 82.4 & 36.6 & 54.3   & 32.1     & 9.6 $\downarrow$ & 76  & 1.8  \\
PyramidDrop~\cite{xing2024pyramiddrop} & 66.7 & 66.9  & 61.1  & 51.2 & 53.6  & 45.1  & 55.4 & 56.2 & 84.4 & 38.9 & 54.5   & 32.3     & 8.9 $\downarrow$ & 81  & 1.87 \\
FEATHER~\cite{endo2024feather}     & 65.2 & 67.2  & 61.3  & 51.9 & 54.3  & 44.7  & 55.9 & 56.5 & 84.8 & 39.2 & 55.1   & 32.4  & 8.7 $\downarrow$ & 85  & 1.92 \\ \midrule
\textbf{LVLM\_CSP} &
  \textbf{70.6} &
  \textbf{73.6} &
  \textbf{66.8} &
  \textbf{56.9} &
  \textbf{61.4} &
  \textbf{50.7} &
  \textbf{62.7} &
  \textbf{64.1} &
  \textbf{96.2} &
  \textbf{46.5} &
  \textbf{58.9} &
  \textbf{34.3} &
  \textbf{2.5} $\downarrow$ &
  \textbf{73} &
  \textbf{1.76} \\ \bottomrule
\end{tabular}
}
\end{table*}

\section{Experiment}
\begin{table}[]
\centering
\caption{Comparison of LVLM\_CSP with previous LVLM visual token pruning methods on LVLM-1.5-7B, using Llama2 7B as the language model and \textbf{CLIP ViT-L patch14-336} as the vision encoder, at a various pruning ratios, evaluated across two reasoning segmentation datasets.}
\label{tab:lisa7b_336}
\resizebox{0.9\linewidth}{!}{%
\begin{tabular}{cccccccc}
\toprule
\multicolumn{1}{c|}{} &
  \multicolumn{3}{c|}{\textbf{RefCOCO}} &
  \multicolumn{2}{c|}{\textbf{RefCOCOg}} &
  $\textbf{N}_\textbf{avg}$ &
  \textbf{TFLOPs} \\ \cline{2-8} 
\multicolumn{1}{c|}{} &
  \textbf{val} &
  \textbf{testA} &
  \multicolumn{1}{c|}{\textbf{testB}} &
  \textbf{val} &
  \multicolumn{1}{c|}{\textbf{test}} &
   &
   \\ \midrule
\multicolumn{1}{c|}{LLaVA-1.5-7B} &
  74.4 &
  77.2 &
  \multicolumn{1}{c|}{69.3} &
  67.9 &
  \multicolumn{1}{c|}{68.6} &
  576 &
  8.2 \\ \midrule
\multicolumn{8}{c}{\textbf{Dropping Ratio 70\%}} \\ \midrule
\multicolumn{1}{c|}{PruMerge} &
  68.9 &
  71.3 &
  \multicolumn{1}{c|}{63.3} &
  62.9 &
  \multicolumn{1}{c|}{63.2} &
  172 &
  3.04 \\
\multicolumn{1}{c|}{VisionZip} &
  70.7 &
  73.3 &
  \multicolumn{1}{c|}{65.3} &
  64 &
  \multicolumn{1}{c|}{63.9} &
  176 &
  3.01 \\
\multicolumn{1}{c|}{FastV} &
  69.9 &
  72.1 &
  \multicolumn{1}{c|}{64.5} &
  62.5 &
  \multicolumn{1}{c|}{62.8} &
  176 &
  3.1 \\
\multicolumn{1}{c|}{PyramidDrop} &
  70.4 &
  72.7 &
  \multicolumn{1}{c|}{65.2} &
  62.9 &
  \multicolumn{1}{c|}{63.3} &
  180 &
  3.14 \\ \midrule
\multicolumn{1}{c|}{\textbf{LVLM\_CSP}} &
  \textbf{73.3} &
  \textbf{75.8} &
  \multicolumn{1}{c|}{\textbf{67}} &
  \textbf{66.5} &
  \multicolumn{1}{c|}{\textbf{67.1}} &
  \textbf{170} &
  \textbf{3.01} \\ \midrule
\multicolumn{8}{c}{\textbf{Dropping Ratio 80\%}} \\ \midrule
\multicolumn{1}{c|}{PruMerge} &
  66.8 &
  68.8 &
  \multicolumn{1}{c|}{62.3} &
  60.1 &
  \multicolumn{1}{c|}{60.6} &
  115 &
  2.3 \\
\multicolumn{1}{c|}{VisionZip} &
  68.3 &
  70.8 &
  \multicolumn{1}{c|}{63.7} &
  61.2 &
  \multicolumn{1}{c|}{61.1} &
  115 &
  2.3 \\
\multicolumn{1}{c|}{FastV} &
  65.4 &
  67.9 &
  \multicolumn{1}{c|}{60.8} &
  57.3 &
  \multicolumn{1}{c|}{56.8} &
  111 &
  2.25 \\
\multicolumn{1}{c|}{PyramidDrop} &
  65.7 &
  68.1 &
  \multicolumn{1}{c|}{61.1} &
  57.5 &
  \multicolumn{1}{c|}{57.1} &
  114 &
  2.29 \\ \midrule
\multicolumn{1}{c|}{\textbf{LVLM\_CSP}} &
  \textbf{70.6} &
  \textbf{73.1} &
  \multicolumn{1}{c|}{\textbf{62.2}} &
  \textbf{62.6} &
  \multicolumn{1}{c|}{\textbf{63.4}} &
  \textbf{115} &
  \textbf{2.3} \\ \midrule
\multicolumn{8}{c}{\textbf{Dropping Ratio 90\%}} \\ \midrule
\multicolumn{1}{c|}{PruMerge} &
  60.9 &
  61.6 &
  \multicolumn{1}{c|}{58.1} &
  53.9 &
  \multicolumn{1}{c|}{53.9} &
  58 &
  1.57 \\
\multicolumn{1}{c|}{VisionZip} &
  64 &
  65.9 &
  \multicolumn{1}{c|}{59.6} &
  55.1 &
  \multicolumn{1}{c|}{55.5} &
  60 &
  1.59 \\
\multicolumn{1}{c|}{FastV} &
  55.4 &
  57.4 &
  \multicolumn{1}{c|}{52.8} &
  47.1 &
  \multicolumn{1}{c|}{46.8} &
  64 &
  1.64 \\
\multicolumn{1}{c|}{PyramidDrop} &
  55.7 &
  58.9 &
  \multicolumn{1}{c|}{54.5} &
  47.7 &
  \multicolumn{1}{c|}{47.9} &
  68 &
  1.7 \\ \midrule
\multicolumn{1}{c|}{\textbf{LVLM\_CSP}} &
  \textbf{65.6} &
  \textbf{67.5} &
  \multicolumn{1}{c|}{\textbf{60.8}} &
  \textbf{57.4} &
  \multicolumn{1}{c|}{\textbf{57.5}} &
  \textbf{66} &
  \textbf{1.67} \\ \bottomrule
\end{tabular}%
}
\end{table}

\begin{table}[]
\centering
\caption{Comparison of LVLM\_CSP with previous LVLM visual token pruning methods on LLaVA-1.5-13B, using Llama2 13B as the language
model and CLIP ViT-L patch14-224 as the vision encoder, at a pruning ratio of approximate 70\%, evaluated across two reasoning segmentation datasets.}
\label{tab:lisa13b_224}
\resizebox{0.9\linewidth}{!}{%
\begin{tabular}{c|cc|cc|cc}
\toprule
\multicolumn{1}{l|}{} & \multicolumn{2}{c|}{\textbf{RefCOCO}} & \multicolumn{2}{c|}{\textbf{RefCOCO+}} & $\textbf{N}_\textbf{avg}$      & \textbf{GFLOPS}      \\ \midrule
\multicolumn{1}{l|}{} & testA             & testB             & testA              & testB             & \multicolumn{1}{l}{} & \multicolumn{1}{l}{} \\ \midrule
LLaVA-1.5-13B~\cite{lai2024lisa}     & 80.1 & 73.6 & 72.8 & 60.1 & 256  & 8.05 \\ \midrule
PruMerge~\cite{shang2024llava}    & 69.2 & 62.9 & 58.8 & 48.6 & 75   & 3.49 \\
VisionZip~\cite{yang2024visionzip}    & 72.8 & 65.8 & 62.1 & 51.4 & 80   & 3.62 \\
FastV~\cite{chen2024image}        & 72.3 & 67   & 61.4 & 51.9 & 73   & 3.45 \\ \midrule
\textbf{LLaVA}\_\textbf{CSP} & 77   & 68.8 & 66.2 & 54.5 & 74.8 & 3.37 \\ \bottomrule
\end{tabular}%
}
\end{table}

\begin{table}[]
\centering
\caption{Ablation study on the clustering module design and the effect of the number of cluster points ($N_c$) on final reasoning segmentation accuracy. We assume $L_c = 8$, $L_s = 6$, and $N_p = 16$.}
\label{tab:lisa7b_ablation1}
\resizebox{0.9\columnwidth}{!}{%
\begin{tabular}{cccccccc}
\toprule
\multicolumn{1}{c|}{} &
  \multicolumn{3}{c|}{\textbf{RefCOCO+}} &
  \multicolumn{2}{c|}{\textbf{RefCOCOg}} &
  $\textbf{N}_\textbf{avg}$ &
  \textbf{TFLOPs} \\ \cline{2-8}
\multicolumn{1}{c|}{} &
  val &
  testA &
  \multicolumn{1}{c|}{testB} &
  val &
  \multicolumn{1}{c|}{test} &
   &
   \\ \midrule
\multicolumn{8}{c}{Nc = 16} \\ \midrule
\multicolumn{1}{c|}{\textbf{Random}} &
  53.5 &
  58.2 &
  \multicolumn{1}{c|}{46.8} &
  59.8 &
  \multicolumn{1}{c|}{61.2} &
  61 &
  1.61 \\
\multicolumn{1}{c|}{\textbf{Uniform}} &
  52.9 &
  57.8 &
  \multicolumn{1}{c|}{47.1} &
  59.5 &
  \multicolumn{1}{c|}{60.9} &
  61 &
  1.61 \\
\multicolumn{1}{c|}{\textbf{CLS}} &
  52.7 &
  57.6 &
  \multicolumn{1}{c|}{47.8} &
  59.3 &
  \multicolumn{1}{c|}{60.8} &
  60 &
  1.59 \\
\multicolumn{1}{c|}{\textbf{Seg First}} &
  \textbf{54.2} &
  \textbf{58.6} &
  \multicolumn{1}{c|}{\textbf{47.9}} &
  \textbf{60.9} &
  \multicolumn{1}{c|}{\textbf{61.9}} &
  60 &
  1.59 \\ \midrule
\multicolumn{8}{c}{Nc = 32} \\ \midrule
\multicolumn{1}{c|}{\textbf{Random}} &
  54.5 &
  59.7 &
  \multicolumn{1}{c|}{48} &
  60.4 &
  \multicolumn{1}{c|}{62.3} &
  65 &
  1.66 \\
\multicolumn{1}{c|}{\textbf{Uniform}} &
  54.7 &
  59.5 &
  \multicolumn{1}{c|}{48.8} &
  61.1 &
  \multicolumn{1}{c|}{62} &
  65 &
  1.66 \\
\multicolumn{1}{c|}{\textbf{CLS}} &
  54.2 &
  58.7 &
  \multicolumn{1}{c|}{\textbf{48.9}} &
  60.4 &
  \multicolumn{1}{c|}{61.7} &
  65 &
  1.66 \\
\multicolumn{1}{c|}{\textbf{Seg First}} &
  \textbf{55.3} &
  \textbf{60.1} &
  \multicolumn{1}{c|}{\textbf{48.9}} &
  \textbf{61.7} &
  \multicolumn{1}{c|}{\textbf{63.3}} &
  65 &
  1.66 \\ \midrule
\multicolumn{8}{c}{Nc = 64} \\ \midrule
\multicolumn{1}{c|}{\textbf{Random}} &
  56.1 &
  60.7 &
  \multicolumn{1}{c|}{49.1} &
  61.8 &
  \multicolumn{1}{c|}{63.2} &
  73 &
  1.76 \\
\multicolumn{1}{c|}{\textbf{Uniform}} &
  56.9 &
  61.3 &
  \multicolumn{1}{c|}{50.6} &
  62.7 &
  \multicolumn{1}{c|}{63.9} &
  73 &
  1.76 \\
\multicolumn{1}{c|}{\textbf{CLS}} &
  55.4 &
  59.5 &
  \multicolumn{1}{c|}{49.8} &
  62.4 &
  \multicolumn{1}{c|}{61.3} &
  74 &
  1.77 \\
\multicolumn{1}{c|}{\textbf{Seg First}} &
  \textbf{57.1} &
  \textbf{61.6} &
  \multicolumn{1}{c|}{\textbf{50.8}} &
  \textbf{63.2} &
  \multicolumn{1}{c|}{\textbf{64.3}} &
  72 &
  1.75 \\ \bottomrule
\end{tabular}%
}
\end{table}

\subsection{Setup}
\noindent \textbf{Datasets} To evaluate our image token reduction framework, we conduct experiments on five reasoning-based segmentation datasets: the RefCOCO series~\cite{yu2016modeling} (RefCOCO, RefCOCO+, RefCOCOg), ReasonSeg~\cite{lai2024lisa}, and RIO~\cite{qu2023rio}. RefCOCO and RIO are based on MS COCO 2014~\cite{lin2014microsoft}, and all datasets require reasoning over a textual query to segment the referred object. The RefCOCO series includes explicit object names and emphasizes semantic and spatial reasoning across multiple instances. In contrast, ReasonSeg and RIO contain task-driven queries that implicitly refer to the target object. We follow standard protocol and report segmentation accuracy using mean Intersection over Union (mIoU). All experiments are run on four NVIDIA RTX A6000 GPUs.


\noindent \textbf{Models} As illustrated in Figure~\ref{fig:task_motivation}(b), state-of-the-art reasoning segmentation methods typically follow a paradigm where an LVLM guides a segmentation model. Following prior work~\cite{lai2024lisa,qian2024affordancellm,rasheed2024glamm,chen2024vltp}, we adopt LLaVA-1.5~\cite{liu2023visual} as the LVLM and ViT-H SAM~\cite{kirillov2023segment} as the segmentation model. To evaluate the scalability of \textbf{LVLM\_CSP}, we experiment with both LLaVA-1.5-7B and 13B variants. Both use CLIP ViT-L~\cite{radford2021learning} as the vision encoder and Llama 2~\cite{touvron2023llama} as the language model. We also assess performance across diffferent input resolutions: LLaVA-1.5-7B is evaluated at 224$\times$224 and 336$\times$336, while the 13B model is tested at 224$\times$224. 


\noindent \textbf{Efficiency Evaluation} Since LVLM\_CSP is a training-free framework, this work focuses on the inference efficiency of the LVLM. Following previous works~\cite{chen2024image,xing2024pyramiddrop}, we report the number of floating-point operations in tera (TFLOPs) for LVLM computations as one of the inference throughput metrics. Additionally, we report the average number of image tokens per language model layer ($N_{avg}$) and LVLM running latency ($T_{LVLM}$) to demonstrate the effectiveness of the visual token reduction framework. For the TFLOPs, we consider the floating-point operations of the multi-head attention and feed-forward network (FFN) modules, calculated as $4nd^2 + 2n^2d + 3ndm$, where $n$ is the total number of tokens, which is the sum of the length of system token, image token, user token, and generated output token, $d$ is the hidden state size, and $m$ is the intermediate size of the FFN. The total number of tokens is computed as:
\begin{equation}
    n = N_{sys} + N_{img} + N_{usr} + N_{out} \quad where \quad N_{img} \in \{N_c, N_s, N_p\}
\end{equation}
Since the three stages of LVLM\_CSP involve different numbers of visual tokens, the total TFLOPs of LVLM\_CSP is computed as:
\begin{equation}
    \sum_{L \in \{L_c, L_s, L_p\}} L \times (4nd^2 + 2n^2d + 3ndm)
\end{equation}

\noindent \textbf{LVLM\_CSP Setup} As shown in Figure~\ref{fig:CSP_arch}, both the reasoning segmentation accuracy and the LVLM acceleration efficiency of LVLM\_CSP are directly influenced by the configuration of its three stages: clustering, scattering, and pruning. Specifically, to reduce the TFLOPs of the LVLM, we aim to minimize the number of scattering layers ($L_s$). During the coarse-grained reasoning stage, we expect the number of clustering tokens ($N_c$) to be significantly smaller than the total number of image tokens ($N$); for example, $N_c$ could be 64 out of a total of 256 image tokens. After fine-grained reasoning, only the most critical image tokens should participate in the remaining computation, thus $N_p$ should be much smaller than $N$—e.g., $N_p$ could be 16 out of 256. To target different image token reduction ratios, we adjust the configuration of the three stages in LVLM\_CSP to achieve optimal reasoning segmentation accuracy. In section~\ref{sec:ablation}, we will examine how each component of the LVLM\_CSP affects both efficiency and accuracy.

Additionally, for the Seg-First clustering module (as shown in Figure~\ref{fig:CSP_arch}(B)(4)), we employ YOLOv11-nano~\cite{redmon2016you}, which requires only 10.4 GFLOPs. Since this is significantly lower than the computational cost of the LLaVA-1.5-7B and LLaVA-1.5-13B, its overhead can be considered negligible. 

\begin{table}[]
\centering
\caption{Ablation study on the the effect of the number of pruning rank ($N_p$) and scattering layer ($L_s$) on final reasoning segmentation accuracy. We assume $L_p = 18$ and $N_c = 64$.}
\label{tab:lisa7b_ablation2}
\resizebox{0.9\columnwidth}{!}{%
\begin{tabular}{cccccccc}
\toprule
\multicolumn{1}{c|}{\textbf{}} & \multicolumn{3}{c|}{\textbf{RefCOCO+}} & \multicolumn{2}{c|}{\textbf{RefCOCOg}} & $\textbf{N}_\textbf{avg}$ & \textbf{TFLOPs} \\ \midrule
\multicolumn{1}{c|}{}               & val  & testA & \multicolumn{1}{c|}{testB} & val  & \multicolumn{1}{c|}{test} &    &      \\ \midrule
\multicolumn{8}{c}{$\textbf{L}_\textbf{c}$\textbf{=6, }$\textbf{L}_\textbf{s}\textbf{=8}$}                                                                                    \\ \hline
\multicolumn{1}{c|}{$\textbf{N}_\textbf{p}$\textbf{=4}}  & 59.1 & 63.9  & \multicolumn{1}{c|}{52.2}  & 64.7 & \multicolumn{1}{c|}{65.9} & 78 & 1.82 \\
\multicolumn{1}{c|}{$\textbf{N}_\textbf{p}$\textbf{=8}}  & 59.3 & 64    & \multicolumn{1}{c|}{52.5}  & 64.9 & \multicolumn{1}{c|}{66}   & 80 & 1.85 \\
\multicolumn{1}{c|}{$\textbf{N}_\textbf{p}$\textbf{=16}} & 59.3 & 64.1  & \multicolumn{1}{c|}{52.5}  & 64.9 & \multicolumn{1}{c|}{66}   & 85 & 1.92 \\
\multicolumn{1}{c|}{$\textbf{N}_\textbf{p}$\textbf{=32}} & 59.3 & 63.8  & \multicolumn{1}{c|}{52.6}  & 65   & \multicolumn{1}{c|}{66}   & 94 & 2.03 \\ \midrule
\multicolumn{8}{c}{$\textbf{L}_\textbf{c}$\textbf{=7, }$\textbf{L}_\textbf{s}\textbf{=7}$}                                                                                    \\ \midrule
\multicolumn{1}{c|}{$\textbf{N}_\textbf{p}$\textbf{=4}}  & 58.3 & 63.3  & \multicolumn{1}{c|}{51.8}  & 64.1 & \multicolumn{1}{c|}{65.3} & 72 & 1.75 \\
\multicolumn{1}{c|}{$\textbf{N}_\textbf{p}$\textbf{=8}}  & 58.7 & 63.3  & \multicolumn{1}{c|}{51.8}  & 64.5 & \multicolumn{1}{c|}{65.4} & 74 & 1.78 \\
\multicolumn{1}{c|}{$\textbf{N}_\textbf{p}$\textbf{=16}} & 58.7 & 63.3  & \multicolumn{1}{c|}{52.1}  & 64.5 & \multicolumn{1}{c|}{65.5} & 79 & 1.84 \\
\multicolumn{1}{c|}{$\textbf{N}_\textbf{p}$\textbf{=32}} & 58.8 & 63.2  & \multicolumn{1}{c|}{52.1}  & 64.7 & \multicolumn{1}{c|}{65.6} & 88 & 1.96 \\ \midrule
\multicolumn{8}{c}{$\textbf{L}_\textbf{c}$\textbf{=8, }$\textbf{L}_\textbf{s}\textbf{=6}$}                                                                                    \\ \midrule
\multicolumn{1}{c|}{$\textbf{N}_\textbf{p}\textbf{=4}$}  & 56.4 & 61.4  & \multicolumn{1}{c|}{50.5}  & 62.8 & \multicolumn{1}{c|}{63.8} & 66 & 1.68 \\
\multicolumn{1}{c|}{$\textbf{N}_\textbf{p}\textbf{=8}$}  & 56.8 & 61.9  & \multicolumn{1}{c|}{50.8}  & 62.5 & \multicolumn{1}{c|}{64.1} & 68 & 1.7  \\
\multicolumn{1}{c|}{$\textbf{N}_\textbf{p}\textbf{=16}$} & 56.9 & 61.8  & \multicolumn{1}{c|}{51}    & 63.1 & \multicolumn{1}{c|}{64.2} & 73 & 1.77 \\
\multicolumn{1}{c|}{$\textbf{N}_\textbf{p}\textbf{=32}$} & 56.9 & 62    & \multicolumn{1}{c|}{50.7}  & 63.4 & \multicolumn{1}{c|}{64.5} & 82 & 1.88 \\ \bottomrule
\end{tabular}%
}
\end{table}

\begin{figure*}[t]
  \centering \includegraphics[width=\textwidth]{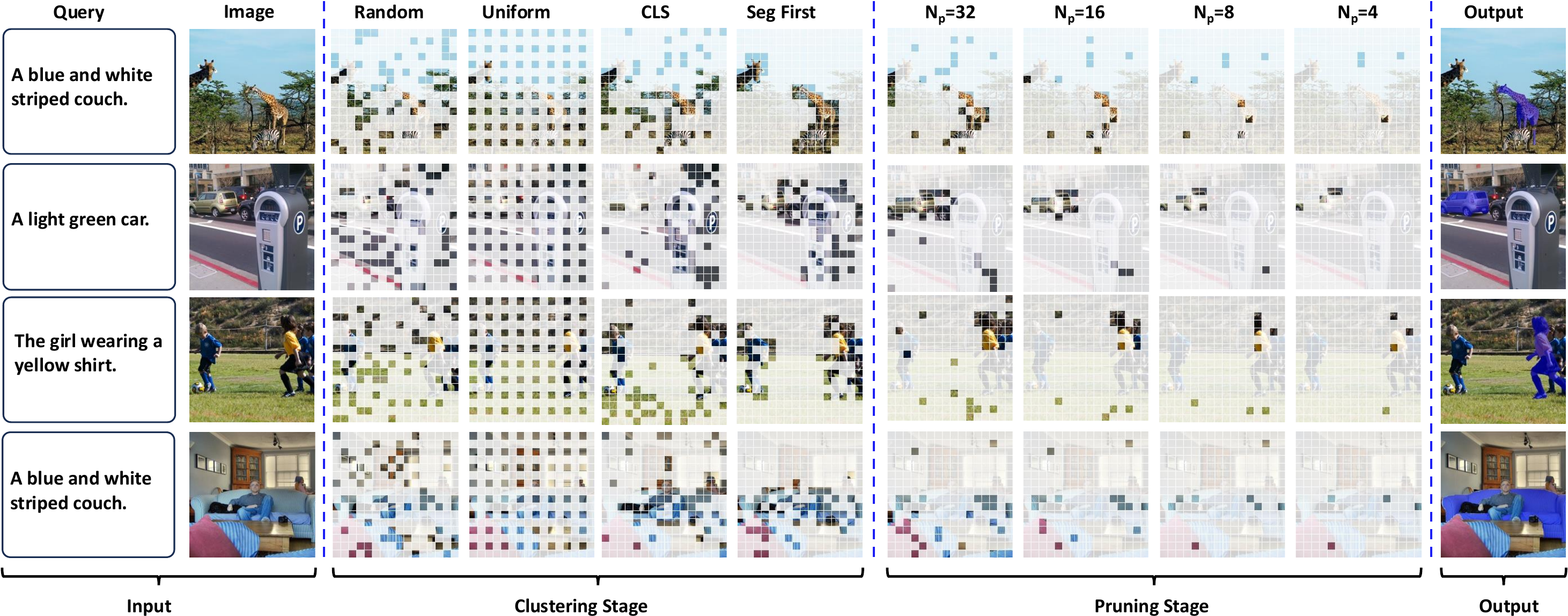}
  \caption{Visualization of \textbf{LVLM\_CSP} token reduction at different stages, along with the input image, query text, and final segmentation masks. For the clustering stage, we show token reduction results from different clustering modules. For the pruning stage, we visualize token reduction under varying values of $N_p$.}
  \label{fig:res_vis}
\end{figure*}

\subsection{Efficiency of LVLM\_CSP}
In this section, we compare \textbf{LVLM\_CSP} with previous LVLM image reduction methods. Instead of restricting the comparison solely to \textbf{FastV}~\cite{chen2024image} and \textbf{PruMerge}~\cite{shang2024llava}, we additionally consider their enhanced versions. Specifically, \textbf{PruMerge+}~\cite{shang2024llava} and \textbf{VisionZip}~\cite{yang2024visionzip} enhance the original PruMerge by keeping extra image tokens for image regions with low attention scores. \textbf{Simignore}~\cite{zhang2024enhancing} and \textbf{SparseVLM}~\cite{zhang2024sparsevlm} improve upon FastV by computing attention scores between image tokens and critical text tokens. On the other hand, \textbf{PyramidDrop}~\cite{xing2024pyramiddrop} and \textbf{FEATHER}~\cite{endo2024feather} aim to preserve more tokens in the early layers of LLaVA’s attention mechanism, while increasing the drop ratio in deeper layers. However, as shown in Table~\ref{tab:lisa7b_224}, Table~\ref{tab:lisa7b_336}, and Table~\ref{tab:lisa13b_224}, none of these prior methods achieve optimal performance on reasoning segmentation tasks when compared to \textbf{LVLM\_CSP}.


In Table~\ref{tab:lisa7b_224}, we compare prior methods with \textbf{LVLM\_CSP} on \textbf{LLaVA-1.5-7B} using a \textbf{CLIP ViT-L} vision encoder and 224$\times$224 image resolution. To ensure fairness, all methods reduce both the average number of image tokens and image-token MHSA TFLOPs by approximately \textbf{70\%}. For \textbf{LVLM\_CSP}, we use ($L_c$, $N_c$, $L_s$, $N_p$) = (8, 64, 6, 16): 8 MHSA layers in the clustering stage with 64 tokens retained, followed by 6 MHSA layers in the scattering stage, and 16 tokens retained in the pruning stage. We omit $N_s$ and $L_p$ since $N_s$ is fixed at 256 (the original number of tokens), and $L_p = 32 - L_c - L_s$ for LLaVA-1.5-7B. Note that this configuration is not optimal. As shown in Section~\ref{sec:ablation}, further tuning improves \textbf{LVLM\_CSP}’s performance on reasoning-based segmentation.

Table~\ref{tab:lisa7b_224} shows that with a 70\% reduction in image tokens and MHSA TFLOPs, \textbf{LVLM\_CSP} outperforms all prior methods by at least \textbf{5\%} in mIoU. \textbf{Previous approaches fall short on reasoning-based segmentation tasks because they focus solely on discarding less-informative tokens without considering the reasoning mechanisms of LVLMs.} Effective spatial and semantic reasoning requires sufficient visual context—i.e., a fine-grained reasoning process—to guide the LVLM toward the correct solution. \textbf{LVLM\_CSP} addresses the high cost of fine-grained reasoning by first applying a lightweight, coarse-grained reasoning stage. For example, in the clustering stage, only 64 out of 256 image tokens are retained, allowing the model to establish a rough task direction. This is followed by a more focused fine-grained reasoning phase—e.g., 6 MHSA layers with full image token participation—to identify key tokens. Finally, just 16 critical tokens are preserved for the remaining MHSA layers. This staged reduction significantly lowers the average number of tokens used during inference, thus reducing TFLOPs while maintaining high segmentation accuracy. The three-stage design—from coarse to fine reasoning—enables both efficiency and precision in reasoning-driven segmentation.

To verify the effectiveness of \textbf{LVLM\_CSP} with higher input image resolutions for the vision encoder and a larger language model decoder, we present additional comparisons in Table~\ref{tab:lisa7b_336} and Table~\ref{tab:lisa13b_224}. Specifically, in Table~\ref{tab:lisa7b_336}, we also compare with previous methods under different token dropping ratios. Due to space limitations, we include only two datasets and four prior methods in this comparison. \textbf{LVLM\_CSP} consistently outperforms all baselines across both \textbf{LLaVA-1.5-7B} with input resolutions of 336$\times$336, as well as \textbf{LLaVA-1.5-13B} with input resolution of 224$\times$224. These results demonstrate that \textbf{LVLM\_CSP} provides significant improvements in model acceleration for reasoning segmentation tasks compared to previous approaches.

\subsection{Ablation Study} \label{sec:ablation}

\begin{table}[]
\centering
\caption{
Performance comparison of LVLM\_CSP across different LVLM models under different hyperparameter settings.
} 
\label{tab:ablation_final}
\resizebox{\linewidth}{!}{%
\begin{threeparttable}
\begin{tabular}{cccccccc}
\toprule
\multicolumn{8}{c}{\textbf{LLaVA-1.5-7B CLIP ViT-L-224x224}}                                                                                  \\ \midrule
\multicolumn{1}{c|}{\textbf{Pruning Ratio (\%)}} & 0     & 65           & 70          & 75          & 80          & 85          & 90          \\ \midrule
\multicolumn{1}{c|}{(\textbf{Lc,Nc,Ls,Np)}}     & -     & (6,96,8,8)   & (7,80,7,8)  & (8,64,6,4)  & (10,48,4,8) & (11,32,3,8) & (12,32,2,2) \\
\multicolumn{1}{c|}{$\textbf{N}_\textbf{avg}$}             & 256   & 86           & 78          & 66          & 51          & 39          & 29          \\
\multicolumn{1}{c|}{\textbf{TFLOPs}}             & 4.12  & 1.93         & 1.83        & 1.68        & 1.48        & 1.33        & 1.2         \\
\multicolumn{1}{c|}{$\textbf{T}_\textbf{LVLM} \textbf{(ms)}^a$}      & 60.39 & 31.28        & 30.02       & 28.88       & 27.5        & 25.45       & 22.8        \\ \midrule
\multicolumn{1}{c|}{$\textbf{mIoU}_\textbf{drop} \textbf{(\%)}^b$}           & 0     & 0.3          & 1.1         & 2.8         & 5           & 7.4         & 8.1         \\
\multicolumn{1}{c|}{$\textbf{mIoU*}_\textbf{drop}$ \textbf{(\%)}}          & 0     & 0.1          & 0.7         & 1.7         & 3.8         & 6.5         & 7.2         \\ \bottomrule \toprule
\multicolumn{8}{c}{\textbf{LLaVA-1.5-7B CLIP ViT-L-336x336}}                                                                                  \\ \midrule
\multicolumn{1}{c|}{\textbf{Pruning Ratio (\%)}} & 0     & 65           & 70          & 75          & 80          & 85          & 90          \\ \midrule
\multicolumn{1}{c|}{\textbf{(Lc,Nc,Ls,Np)}}     & -     & (6,128,9,32) & (6,32,9,16) & (7,64,7,16) & (8,24,6,8)  & (9,32,4,8)  & (11,32,3,2) \\
\multicolumn{1}{c|}{$\textbf{N}_\textbf{avg}$}             & 576   & 203          & 176         & 149         & 118         & 85          & 66          \\
\multicolumn{1}{c|}{\textbf{TFLOPs}}             & 8.25  & 3.44         & 3.1         & 2.74        & 2.35        & 1.92        & 1.68        \\
\multicolumn{1}{c|}{$\textbf{T}_\textbf{LVLM}$ \textbf{(ms)}}      & 123.5 & 58.3         & 55.33       & 50.46       & 45.3        & 39.16       & 35.16       \\ \midrule
\multicolumn{1}{c|}{$\textbf{mIoU}_\textbf{drop}$ \textbf{(\%)}}           & 0     & 0.3          & 1           & 1.4         & 4.4         & 7.5         & 8.5         \\
\multicolumn{1}{c|}{$\textbf{mIoU*}_\textbf{drop}$ \textbf{(\%)}}          & 0     & 0.11         & 0.6         & 0.9         & 3.1         & 6.5         & 7.4         \\ \bottomrule \toprule
\multicolumn{8}{c}{\textbf{LLaVA-1.5-13B CLIP ViT-L-224x224}}                                                                                 \\ \midrule
\multicolumn{1}{c|}{\textbf{Pruning Ratio (\%)}} & 0     & 65           & 70          & 75          & 80          & 85          & 90          \\ \midrule
\multicolumn{1}{c|}{\textbf{(Lc,Nc,Ls,Np)}}     & -     & (5,64,11,16) & (6,64,10,8) & (7,32,9,8)  & (9,16,7,8)  & (11,16,5,8) & (12,32,2,8) \\
\multicolumn{1}{c|}{$\textbf{N}_\textbf{avg}$}             & 256   & 88           & 78          & 68          & 53          & 41          & 27          \\
\multicolumn{1}{c|}{\textbf{TFLOPs}}             & 8.05  & 3.83         & 3.58        & 3.32        & 2.95        & 2.65        & 2.31        \\
\multicolumn{1}{c|}{$\textbf{T}_\textbf{LVLM}$ \textbf{(ms)}}      & 104   & 54.5         & 52.63       & 49.8        & 45.2        & 42          & 40.08       \\ \midrule
\multicolumn{1}{c|}{$\textbf{mIoU}_\textbf{drop} \textbf{(\%)}$}           & 0     & 1.9          & 2.5         & 3.6         & 6.6         & 9.7         & 11.1        \\
\multicolumn{1}{c|}{$\textbf{mIoU*}_\textbf{drop}$ \textbf{(\%)}}          & 0     & 1.1          & 1.2         & 2.4         & 5.1         & 8.6         & 10.2        \\ \bottomrule
\end{tabular}%
\begin{tablenotes}[leftmargin=0.35cm, itemindent=.0cm, itemsep=0.0cm, topsep=0.1cm]
\item[a] $T_{\text{LVLM}}$ denotes the average latency for LVLM reasoning over one text and one image on a RTX A6000. 
\item[b] $\boldsymbol{\mathrm{mIoU}_{\mathrm{drop}}}$ represents the average mIoU accuracy drop across five reasoning segmentation datasets compared to the original LVLM. $\boldsymbol{\mathrm{mIoU}_{\mathrm{drop}}}$ reflects the training-free accuracy, while $\boldsymbol{\mathrm{mIoU^*}_{\mathrm{drop}}}$ indicates the accuracy after fine-tuning the segmentation model, with the LVLM kept frozen.
\end{tablenotes}
\end{threeparttable}
}
\end{table}

In this section, we analyze the importance of each component in \textbf{LVLM\_CSP} to optimize reasoning segmentation accuracy under different token dropping ratios. In Table~\ref{tab:lisa7b_ablation1} and Table~\ref{tab:lisa7b_ablation2}, we present experiments conducted using LLaVA-1.5-7B with an input resolution of 224$\times$224. In Figure~\ref{tab:lisa7b_ablation1}, we fix the number of layers in the three stages ($L_c$, $L_s$, and $L_p$) and set the number of retained tokens in the pruning stage to $N_P = 16$. We then focus on how the number of tokens selected during the clustering stage ($N_c$), as well as the design of the clustering module, affect the final reasoning segmentation accuracy. As shown in Figure~\ref{tab:lisa7b_ablation1}, \textit{Seg First} consistently achieves the best performance across all values of $N_c$. When $N_c$ is small, \textit{Random} outperforms both \textit{Uniform} and \textit{CLS}, whereas when $N_c$ is large, \textit{Uniform} achieves performance comparable to \textit{Seg First}. The superior performance of \textit{Seg First} can be attributed to its instance-aware sampling, which helps the LVLM focus on relevant objects and more easily establish the overall reasoning direction. \textit{Uniform} performs well when $N_c$ is large because it selects representative tokens across the entire image. However, with a low $N_c$, it struggles to maintain the same effectiveness due to the inclusion of less informative background tokens. Table~\ref{tab:lisa7b_ablation1} also shows that increasing $N_c$ slightly improves the final reasoning segmentation accuracy. However, a larger $N_c$ also leads to an increase in the average number of image tokens ($N_{\text{avg}}$) and computational cost (in terms of TFLOPs).

Table~\ref{tab:lisa7b_ablation2} highlights the importance of the scattering stage depth ($L_s$) and the number of tokens retained in the pruning stage ($N_p$). The results show that increasing $L_s$ significantly improves reasoning segmentation accuracy. However, this also leads to a rise in the average number of image tokens ($N_{\text{avg}}$) and computational cost (TFLOPs). Additionally, Table~\ref{tab:lisa7b_ablation2} shows that the impact of $N_p$ on final segmentation accuracy is minimal. This suggests that after sufficient coarse-grained and fine-grained reasoning, the LVLM is already capable of attending to the most relevant image tokens needed to resolve the query. Based on the ablation studies in Tables~\ref{tab:lisa7b_ablation1} and~\ref{tab:lisa7b_ablation2}, we conclude that to mitigate the high computational overhead of the fine-grained reasoning stage (scattering), it is effective to increase both the number of layers ($L_c$) and the number of retained tokens ($N_c$) in the clustering stage, while maintaining a very low number of tokens ($N_p$) in the pruning stage and decrease the scattering layers ($N_s$).

Based on the above ablation studies, Table~\ref{tab:ablation_final} presents the optimal configuration of \textbf{LVLM\_CSP} for three different LVLMs, with image token reduction ratios ranging from 65\% to 90\%. We also report the LVLM execution latency ($\text{T}_{\text{LVLM}}$) measured on a single NVIDIA A6000 GPU. To ensure a fair comparison, here we only consider the inference latency with single query image and text. For the average mIoU drop, we report both training-free and fine-tuned results. In the fine-tuning setup, both the LLaVA components and SAM ViT backbone are frozen while only the SAM mask decoder is fine-tuned. Here we freeze the LLaVA to keep the LLM's generality. As shown in Table~\ref{tab:ablation_final}, under the training-free setting, \textbf{LVLM\_CSP} achieves a 65\% reduction in image tokens and TFLOPs with virtually no drop in accuracy. Furthermore, after fine-tuning the SAM mask decoder, \textbf{LVLM\_CSP} can reduce up to 75\% of image tokens and TFLOPs while still maintaining high segmentation accuracy. Upon implementing LVLM\_CSP, we achieve a speedup in LVLM inference ranging from $\times$2 to $\times$4. In Figure~\ref{fig:res_vis}, we visualize the token reduction at each stage of \textbf{LVLM\_CSP}, along with the final reasoning-based segmentation. Figure~\ref{fig:res_vis} shows that during the clustering stage, very few objects have image tokens participating in the LLM MHSA computation. At the pruning stage, only the critical image tokens relevant to answering the query participate in the LLM MHSA computation.

\section{Related Works}
\label{sec:realated works}

\subsection{LVLM and Reasoning Segmentation}
Large vision language models (LVLMs) like LLaVA~\cite{liu2023visual}, MiniGPT-4~\cite{zhu2023minigpt}, and BLIP2~\cite{li2023blip} have shown strong capabilities in visual reasoning~\cite{wang2024exploring} and multimodal understanding~\cite{zhang2024mm}. Recent efforts, including LISA~\cite{lai2024lisa}, GroundHOG~\cite{zhang2024groundhog}, and AffordanceLLM~\cite{qian2024affordancellm}, integrate LVLMs with segmentation models like SAM~\cite{kirillov2023segment} and GroundingDINO~\cite{liu2024grounding} to address reasoning-driven segmentation tasks~\cite{yu2016modeling,lai2024lisa,qu2023rio}. These require both semantic and fine-grained spatial reasoning to generate object masks from text. Compared to models like MDETR~\cite{kamath2021mdetr}, Polyformer~\cite{liu2023polyformer}, and CLIP~\cite{radford2021learning}, LVLMs leverage large-scale multimodal training for superior performance, albeit with higher computational costs. This motivates efficient model compression for reasoning-based segmentation.

\subsection{LVLM Token Pruning}
Token reduction is key to accelerating transformer-based models due to the quadratic cost of MHSA~\cite{tang2024survey}. Prior work has explored pruning in Vision Transformers~\cite{lu2023content,tang2023dynamic,liu2024revisiting}, KV cache compression~\cite{han2023lm,ge2023model}, and LLMs~\cite{liu2023scissorhands}. More recently, LVLM-specific image token reduction~\cite{chen2024image,endo2024feather,zhang2024enhancing,shang2024llava} has focused on high-level understanding tasks. However, reasoning-based segmentation poses additional challenges, requiring finer-grained pruning that preserves token-level details critical for accurate mask generation.

\section{Conclusion}
In this work, we propose a method for reducing image tokens in large vision-language models (LVLMs) when handling reasoning-based segmentation tasks. Our training-free, three-stage framework guides the LVLM through a coarse-to-fine reasoning process, enabling it to maintain a low average number of image tokens involved in computation. Extensive experiments demonstrate that our approach significantly improves computational efficiency while preserving state-of-the-art performance in reasoning-based segmentation.

\section*{Acknowledgements}
This work was supported in part by the DARPA Young Faculty Award, the National Science Foundation (NSF) under Grants \#2127780, \#2319198, \#2321840, \#2312517, and \#2235472, the Semiconductor Research Corporation (SRC), the Office of Naval Research through the Young Investigator Program Award, the U.S. Army Combat Capabilities Development Command (DEVCOM) Army Research Laboratory under Support Agreement No. USMA 21050, and Grants \#N00014-21-1-2225 and N00014-22-1-2067. Additionally, support was provided by the Air Force Office of Scientific Research under Award \#FA9550-22-1-0253, along with generous gifts from Xilinx and Cisco.

\bibliographystyle{ACM-Reference-Format}
\bibliography{sample-base}










\end{document}